\documentclass[11pt]{article}
\usepackage{ijcnlp2011}
\usepackage{times}
\usepackage{latexsym}
\usepackage{amsmath}
\usepackage{multirow}
\usepackage{url}

\usepackage{algorithm}
\usepackage{algorithmic}
\usepackage{amsfonts}
\usepackage{amssymb}
\usepackage{graphics}
\usepackage{graphicx}
\usepackage{hyperref}
\usepackage{mathrsfs}
\usepackage{subfigure}

\title{Inter-rater Agreement on Sentence Formality}

\author{
  Shibamouli Lahiri \\
  Computer Science and Engineering \\
  The Pennsylvania State University \\
  University Park, PA 16802, USA \\
  {\tt shibamouli@psu.edu} \\\And
  Xiaofei Lu \\
  Applied Linguistics \\
  The Pennsylvania State University \\
  University Park, PA 16802, USA \\
  {\tt xxl13@psu.edu} \\
}

\date{}

\begin{document}

\maketitle

\begin{abstract}

Formality is one of the most important dimensions of writing style variation. In this study we conducted an inter-rater reliability experiment for assessing sentence formality on a five-point Likert scale, and obtained good agreement results as well as different rating distributions for different sentence categories. We also performed a difficulty analysis to identify the bottlenecks of our rating procedure. Our main objective is to design an automatic scoring mechanism for sentence-level formality, and this study is important for that purpose.

\end{abstract}

\section{Introduction}
\label{sec:intro}

Formality of language is an important dimension of writing style variation~\cite{biber1988variation,Heylighen99formalityof}. Academic papers are usually written more formally than blog posts, while blog posts are usually written more formally than forum threads~\cite{DBLP:conf/cicling/LahiriML11}. The concept of formality has so far been explored from three different levels - the document-level~\cite{Heylighen99formalityof}, the word-level~\cite{Brooke3}, and the sentence-level~\cite{DBLP:conf/cicling/LahiriML11}. All these studies have directly or indirectly shown that formality is a rather subjective concept, and there exists a \emph{continuum of formality} so that linguistic units (e.g., a word, a sentence or a document) may never be classified as ``fully formal'' or ``fully informal'', but they should rather be \emph{rated on a scale} of formality. For example, consider the following three sentences: ``Howdy!!'', ``How r u?'' and ``How are you?''. Note that each sentence is more formal than the previous one, and the formalization process can be continued forever. Heylighen and Dewaele~\shortcite{Heylighen99formalityof} in their seminal work on document formality have explained this issue by defining two different variants of formality - surface and deep. The surface variant formalizes language for no specific purpose other than stylistic embellishment, but the deep variant formalizes language for communicating the meaning more clearly and completely. More complete communication of meaning involves context-addition, which can be continued ad infinitum, thereby resulting in sentences that are always more (deeply) formal than the last one. Heylighen and Dewaele also discussed the use of formality to obscure meaning (e.g., by politicians), but it was treated as a corruption of the original usage.

Heylighen and Dewaele's quantification of deep formality is not as reliable when we look into the sub-document level. At the word level, a very different approach for dealing with the issue of formality has been proposed by Brooke, et al~\shortcite{Brooke3}. They experimented with several word-level formality scores to determine the one that best associated with hand-crafted seed sets of formal and informal words, as well as words co-occurring with the seed sets. Lahiri, et al.~\shortcite{DBLP:conf/cicling/LahiriML11} explored the concept of sentence-level formality from two different perspectives - deep formality of annotated and un-annotated sentence corpora, and inherent agreement between two judges on an annotation task. They observed that the deep formality of sentences broadly followed the corpus-level trend, and correlated well with human annotation. It was also reported that when the annotation judgment was binary (i.e., formal vs informal sentence) and no prior instructions were given to the annotators as to what constitutes a formal sentence, there was very poor inter-annotator agreement, which in turn showed how inherently subjective the concept of formality is.

Our work is a direct extension of the inter-annotator agreement reported by Lahiri, et al~\shortcite{DBLP:conf/cicling/LahiriML11}. Instead of binary annotation (formal/informal sentence), we adopted a 1-5 Likert scale, where 1 represents a very informal sentence and 5 a very formal sentence. Keeping prior instructions to a minimum, we observed that the inherent agreement results using Likert scale were better than the results using binary annotation. This observation validates the presence of formality continuum at the sentence level. It also helped us construct a seed set of sentences with human-assigned formality ratings. This seed set can be used in evaluating an automatic scoring mechanism for sentence-level formality. Note that adding up word-level scores is not appropriate for this purpose, because it may so happen that all the words in a sentence are formal, but the sentence as a whole is not so formal (e.g., ``For all the stars in the sky, I do not care.'').

This paper is organized as follows. In Section~\ref{sec:expdesign} we explain the design of our study and its rationale. Section~\ref{sec:results} gives the experimental results and difficulty analysis. We conclude in Section~\ref{sec:conclusion}, outlining our contributions.

\section{Study Design}
\label{sec:expdesign}

We adopted a five-point Likert scale for the formality annotation of sentences. The 1-5 scale is easily interpretable, widely used and well-suited for ordinal ratings. The annotators were requested to rate each sentence as follows: 1 - Very Informal, 2 - Informal, 3 - In-between, 4 - Formal, 5 - Very Formal, X - Not Sure. The annotators were not given any instructions as to what constitutes a very formal sentence, what constitutes a very informal sentence, etc. They were, however, advised to keep in mind that the ratings were relative to each other, and were requested to be consistent in their ratings, and rate sentences independently.

We conducted the inter-rater agreement study in two phases. In the warm-up (pilot) phase, we gave 100 sentences to the raters, and observed if they were able to do the ratings on their own, and if the agreement was good or not. Then we proceeded to the actual annotation phase with 500 sentences. The difference between these two phases was that in the warm-up phase, the raters sat together in our presence, working independently and getting a feel of the task. We, however, did not provide any instructions on how to rate the sentences, and the raters were completely on their own. In the actual phase, the raters worked separately and in our absence.

Two raters participated in this study. Both were female undergraduate sophomore students, and both were native English speakers at least 18 years of age. The raters were selected randomly from a pool of respondents who emailed us their consent to participate in this study. The warm-up phase took less than an hour, and the actual phase took approximately one and a half hours.



The sentences were selected from the four datasets used in~\cite{DBLP:conf/cicling/LahiriML11}. For the warm-up set, we randomly picked 25 sentences from each category (blog, news, forum and paper), and for the actual set, we randomly picked 125 sentences from each category. The warm-up set and the actual set were mutually exclusive, and sentences in each set were scrambled so that (a) raters did not know which sentence falls under which category, and (b) raters were not influenced by the original ordering of sentences.

\section{Results}
\label{sec:results}

We performed three types of analysis on the warm-up as well as on the actual set of sentence ratings\footnote{Code and data available at \url{http://www.4shared.com/zip/4_ZicXU2/iaa_sentence_formality_code_an.html}}. The first type attempts to find out the agreement and correlation between the two raters, and how similar the ratings were. The second type of analysis explores the properties of rating distributions and whether distributions for different categories of sentences (i.e., blog, forum, news or paper) are different. The third type of analysis deals with two kinds of difficult sentences and their relative frequencies. The two kinds of difficult sentences are X-marked sentences and sentences for which the raters differed by two or more points.

\subsection{Agreement and Correlation}
\label{subsec:AgreementCorrelation}

\begin{table*}
\begin{center}
\begin{tabular}{cccccc}
\hline
& \textbf{Forum} & \textbf{Blog} & \textbf{News} & \textbf{Paper} & \textbf{Overall}\\
\hline \\
$\gamma$-test & 0.5729 & 0.6509 & 0.6105 & 0.2249 & 0.7212 \\
$\tau_{a}$ & 0.3472 & 0.4580 & 0.3849 & 0.1471 & 0.5406 \\
$\tau_{b}$ & 0.2993 & 0.4228 & 0.3428 & 0.1343 & 0.5053 \\
Spearman's $\rho$ & 0.3818 & 0.5271 & 0.4208 & 0.1625 & 0.6194 \\
\hline \\
Krippendorff's $\alpha$ & 0.3584 & 0.4574 & 0.4022 & 0.0772 & 0.5802 \\
Cohen's $\kappa$ & 0.1874 & 0.1721 & 0.1607 & 0.0542 & 0.2290 \\
\hline \\
Cosine Similarity & 0.9328 & 0.9549 & 0.9725 & 0.9666 & 0.9606 \\
Tanimoto Similarity & 0.8676 & 0.9029 & 0.9455 & 0.9210 & 0.9170 \\
\hline
\end{tabular}
\end{center}
\caption{\label{tab:actual_similarity}Agreement and correlation values on the actual set.}
\end{table*}


We report four nonparametric correlation coefficients between the two raters, as well as cosine and Tanimoto similarity~\cite{tanimoto} between the two rating vectors.\footnote{We used MATLAB for all our analyses.} Each element in a rating vector corresponds to a sentence and the value of the element is the formality rating of the sentence. We also report Cohen's $\kappa$~\cite{cohenkappa} and Krippendorff's $\alpha$~\cite{Krippendorff2007} for measuring quantitative agreement between the two raters. These results were obtained after pruning the X-marked sentences. Table~\ref{tab:actual_similarity} shows the results for the actual set. Overall results (the rightmost column) indicate that the cosine and Tanimoto similarity between the raters were fairly high, which shows that the rating directions were preserved. In other words, if rater A rated sentence S1 as more formal than sentence S2, then rater B also rated S1 as more formal than S2, not the other way round. This shows the consistency of our raters and the importance of Likert scale in formality judgment. High similarity values were also obtained within specific categories (forum, blog, news and paper sentences), showing that rating consistency was maintained across categories. Similar results were obtained for the warm-up set as well.

Correlation between two raters was measured with four non-parametric tests - the $\gamma$-test~\cite{gamma}, Kendall's $\tau_{a}$ and $\tau_{b}$~\cite{kendalltau}, and Spearman's $\rho$. The $\gamma$-test and $\tau_{b}$ are particularly well-suited for measuring similarity between ordinal ratings, because they emphasize the number of concordant pairs over the number of discordant pairs. We obtained a fairly high value for the overall $\gamma$ for both the actual and the warm-up set, thereby showing good inherent agreement between annotators. Values for Kendall's $\tau_{a}$ and $\tau_{b}$, and Spearman's $\rho$ were not as high, but they were all found to be statistically significant (i.e., significantly different from 0) with p-value $<$ 0.05. Only for the ``paper'' category, the p-values were found to be $>$ 0.05 for $\gamma$, Spearman's $\rho$, and Kendall's $\tau_{a}$. For the warm-up set, p-values were found to be $>$ 0.05 for Spearman's $\rho$ and Kendall's $\tau_{a}$ under the ``blog'' category. All others were statistically significant. Note that the p-values for Kendall's $\tau_{b}$, Krippendorff's $\alpha$ and $\gamma$-test are one-tailed and computed by bootstrapping (1000 bootstrap samples) under the null hypothesis that the observed correlation is 0.

Inter-rater reliability was measured with Cohen's $\kappa$ and Krippendorff's $\alpha$. Justification for using the latter is given in~\cite{Artstein:2008:IAC:1479202.1479206}. When category labels are not equally distinct from one another (as is our case), Krippendorff's $\alpha$ must be computed. The values are reported in Table~\ref{tab:actual_similarity}. Note that Krippendorff's $\alpha$ allows missing data as well, so we could have incorporated the X-marked sentences in $\alpha$-computation. But to avoid complication, we chose not to do so, and quarantined the X-marked sentences for further analysis. Observe from Table~\ref{tab:actual_similarity} that although the category-wise $\kappa$-values indicate slight or no agreement~\cite{landiskoch}, the overall $\kappa$-value for the actual set indicates fair agreement. This is a significant achievement given the conservativeness of $\kappa$, the subjectivity associated with formality judgment, our small dataset, and no prior instructions on what to consider formal and what to consider informal. This result is better than the one reported in~\cite{DBLP:conf/cicling/LahiriML11} ($\kappa_{Blog}$ 0.164, $\kappa_{News}$ 0.019), which shows the merit of Likert-scale annotation for formality judgment. The overall $\kappa$-values were found to be statistically significant with p-value $<$ 0.005.

\subsection{Rating Distributions}
\label{subsec:RatingDistribution}

\begin{figure}
\begin{center}
\subfigure[Rater 1 (3.1, 1.05)]
{
\includegraphics[width=0.46\linewidth]{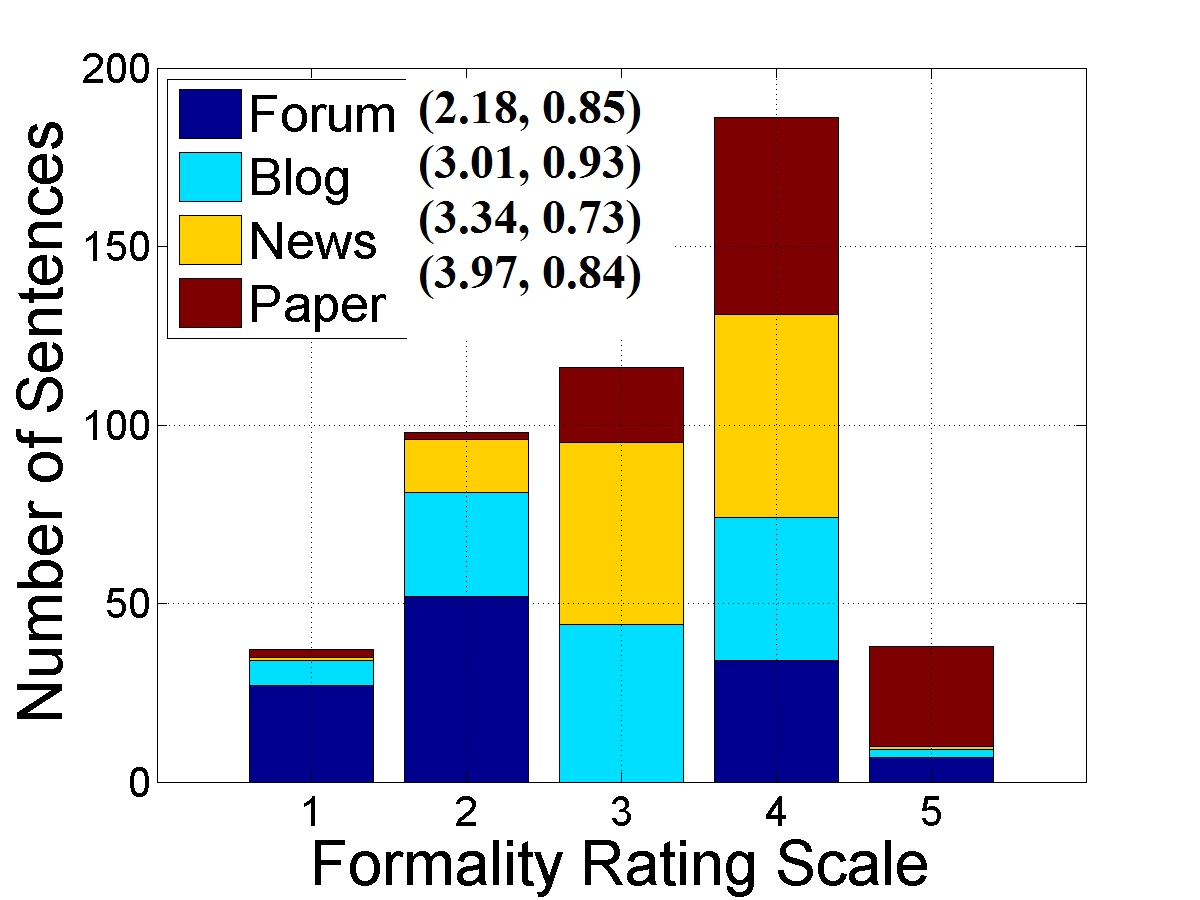}
\label{fig:sentHistRater1}
}
\subfigure[Rater 2 (2.85, 0.92)]
{
\includegraphics[width=0.46\linewidth]{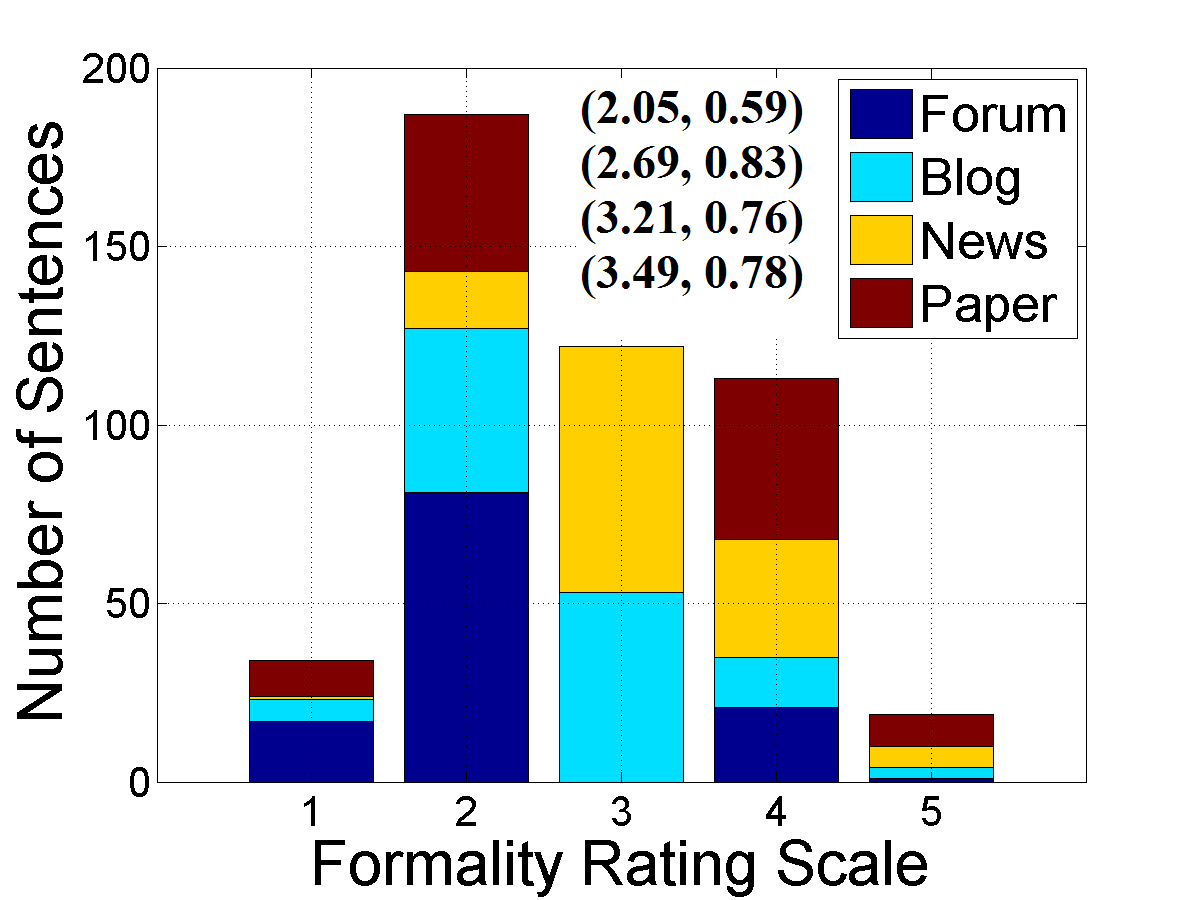}
\label{fig:sentHistRater2}
}
\end{center}
\caption[]{Sentence formality rating histograms for the actual set (mean rating and sd in parentheses).}
\label{fig:sentHist}
\end{figure}

The distributions of sentence formality ratings (Figure~\ref{fig:sentHist}) for the actual set indicate that Rater 1 tended to rate sentences more formally on average than Rater 2 (same conclusion from paired t-test and U test~\cite{mannwhitney} with 95\% confidence). Figure~\ref{fig:sentHist} shows that the two raters rated almost the same number of sentences as either 1 or 3. In other words, the number of very informal as well as ``in-between-type'' sentences appears to be consistent across two raters. But Rater 1 considered a large number of sentences ``formal'' (i.e., rating 4), whereas Rater 2 considered an almost equally large number of sentences informal (i.e., rating 2). On the other hand, relatively fewer sentences were considered ``very formal'' or ``very informal''. One possible reason for this behavior is the so-called ``central tendency bias''\footnote{See, for example, \url{http://en.wikipedia.org/wiki/Likert\_scale}}, which we consider a limitation of our study.

To determine if the rating distributions under different categories (blog, forum, news and paper) were significantly different from each other, we performed the non-parametric Kruskal-Wallis test~\cite{kruskalwallis}. For both raters and for both actual and warm-up sets, the results indicated that at least one category differed from others in formality rating. The non-parametric U Test on category pairs (with Bonferroni correction~\cite{dunn} for multiple comparison) showed the formality ratings under each category to be significantly different from others (95\% confidence). Only in the warm-up set, the blog and news ratings were not found to be significantly different for either of the raters. We also performed a Kolmogorov-Smirnov test~\cite{smirnov} to see if the distributions were significantly different from each other. For the warm-up set, the results followed U Test, although for one rater, blog and forum sentence ratings were not found to be significantly different. For the actual set, for one rater blog and news sentence ratings were not found to be significantly different.

Following the U Test results, we note that the category-wise sentence formality rating distributions were significantly different from each other, and the general trend of mean and median ratings followed the intuition that the ``paper'' category sentences are more formal than the ``blog'' and ``news'' categories, which in turn are more formal than the ``forum'' category.

\subsection{Difficulty Analysis}
\label{subsec:ErrorAnalysis}

There were 25 X-marked sentences in the actual set (5\%), and six in the warm-up set (6\%). These sentences represent confusing cases that at least one rater marked as ``X''. These are primarily system error and warning messages, programming language statements, incomplete sentences, and two sentences merged into one. The last two types of sentences arose because of imprecise sentence segmentation. A manual cleaning to remove such cases from the original datasets seemed prohibitively time-consuming. Many of these sentences are from the ``paper'' category.

The second type of difficulty concerns the sentences for which the annotators differed by two or more points. There were 40 such cases in the actual set, and 7 cases in the warm-up set. These sentences were either too long, or too short, or grammatically inconsistent. Many of them were incomplete sentences, or two sentences merged into one. Note that since we did not provide the annotators with a detailed guideline on what to consider formal and what informal, they freely interpreted the too-long, too-short and grammatically inconsistent sentences according to their own formality judgment. This is precisely where the subjectivity in their judgments kicked in. However, such cases were never a majority.

\section{Conclusion}
\label{sec:conclusion}

In this paper we reported an inter-rater agreement study for assessing sentence formality on a five-point Likert scale. We obtained better and consistent agreement values on a set of 500 sentences. Sentences from different categories (blog, forum, news and paper) were shown to follow different formality rating distributions. We also performed a difficulty analysis to identify problematic sentences, and as a by-product of our study, we obtained a seed set of human-annotated sentences that can later be used in evaluating an automatic scoring mechanism for sentence-level formality.


\end{document}